\documentclass[letterpaper]{article} 
\usepackage[preprint]{neurips_2022}
\usepackage{times}  
\usepackage{helvet}  
\usepackage{courier}  
\usepackage[hyphens]{url}  
\usepackage{graphicx} 
\usepackage{natbib}  
\usepackage{caption} 
\usepackage{algorithm}
\usepackage{algorithmic}

\usepackage{newfloat}
\usepackage{listings}
\DeclareCaptionStyle{ruled}{labelfont=normalfont,labelsep=colon,strut=off} 
\floatstyle{ruled}
\newfloat{listing}{tb}{lst}{}
\floatname{listing}{Listing}
\usepackage{bm}

\title{It's DONE: Direct ONE-shot learning with quantile weight imprinting}

\author{ 
  Kazufumi Hosoda\thanks{Center for Information and Neural Networks (CiNet), Advanced ICT Research Institute, National Institute of Information and Communications Technology (NICT), 1-4 Yamadaoka, Suita, Osaka 565-0871, Japan}, 
\, Keigo Nishida\thanks{Laboratory for Computational Molecular Design, RIKEN Center for Biosystems Dynamics Research (BDR), 6-2-3 Furuedai, Suita, Osaka, 565-0874, Japan}, 
\, Shigeto Seno \thanks{Graduate School of Information Science and Technology, Osaka University}, 
\, Tomohiro Mashita \thanks{Cybermedia Center, Osaka University}
\AND
Hideki Kashioka\footnotemark[1],
\, Izumi Ohzawa\footnotemark[1]
\AND
GitHub: \url{https://github.com/hosodakazufumi/tfdone}
}

\usepackage{bibentry}

\begin{document}

\maketitle

\begin{abstract}
Learning a new concept from one example is a superior function of the human brain and it is drawing attention in the field of machine learning as a one-shot learning task. In this paper, we propose one of the simplest methods for this task with a nonparametric weight imprinting, named Direct ONE-shot learning (DONE). DONE adds new classes to a pretrained deep neural network (DNN) classifier with neither training optimization nor pretrained-DNN modification. DONE is inspired by Hebbian theory and directly uses the neural activity input of the final dense layer obtained from data that belongs to the new additional class as the synaptic weight with a newly-provided-output neuron for the new class, transforming all statistical properties of the neural activity into those of synaptic weight by quantile normalization. DONE requires just one inference for learning a new concept and its procedure is simple, deterministic, not requiring parameter tuning and hyperparameters. DONE overcomes a severe problem of existing weight imprinting methods that DNN-dependently interfere with the classification of original-class images. The performance of DONE depends entirely on the pretrained DNN model used as a backbone model, and we confirmed that DONE with current well-trained backbone models perform at a decent accuracy. 

\end{abstract}

\section{Introduction}

As is well known, artificial neural networks are initially inspired by the biological neural network in the animal brain. Subsequently, Deep Neural Networks (DNNs) achieved great success in computer vision \cite{simonyan_very_2015,he_deep_2016} and other machine learning fields. However, there are lots of tasks that are easy for humans but difficult for current DNNs. One-shot learning is considered as one of those kinds of tasks \cite{fei-fei_one-shot_2006,lake_human-level_2015}. Humans can add a new class to their large knowledge from only one input image but it is difficult for DNNs unless another specific optimization is added. Usually, additional optimizations require extra user skills and calculation costs for tuning parameters and hyperparameters. Thus, for example, if an ImageNet model \cite{deng_imagenet_2009,russakovsky_imagenet_2015} that learned 1000 classes can learn a new class ``baby'' from one image of a baby with neither additional training optimization nor pretrained-DNN modification, it will be useful in actual machine learning uses.

For a DNN model trained with a sufficiently rich set of images, a reasonable representation of unknown images must exist somewhere in the hidden multi-dimensional space. 
Indeed, \textit{weight imprinting}, proposed by Qi et al. \cite{qi_low-shot_2018}, can add novel classes to Convolutional Neural Networks (CNNs) by using final-dense-layer input of a new-class image without extra training. Qi's weight imprinting method needs just a few CNN-architecture modifications and can provide decent accuracy in a one-shot image classification task (e.g., accuracy for novel-class images was 21\% when novel 100 classes were added to the original 100 classes in CUB-200-2011 dataset). Moreover, some studies show that the capabilities of DNN itself have the potential to enable Out-of-Distribution Detection (OOD) \cite{lakshminarayanan_simple_2017,fort_exploring_2021}.

\begin{figure*}[t]
  \centering
  \includegraphics[width=0.95\textwidth]{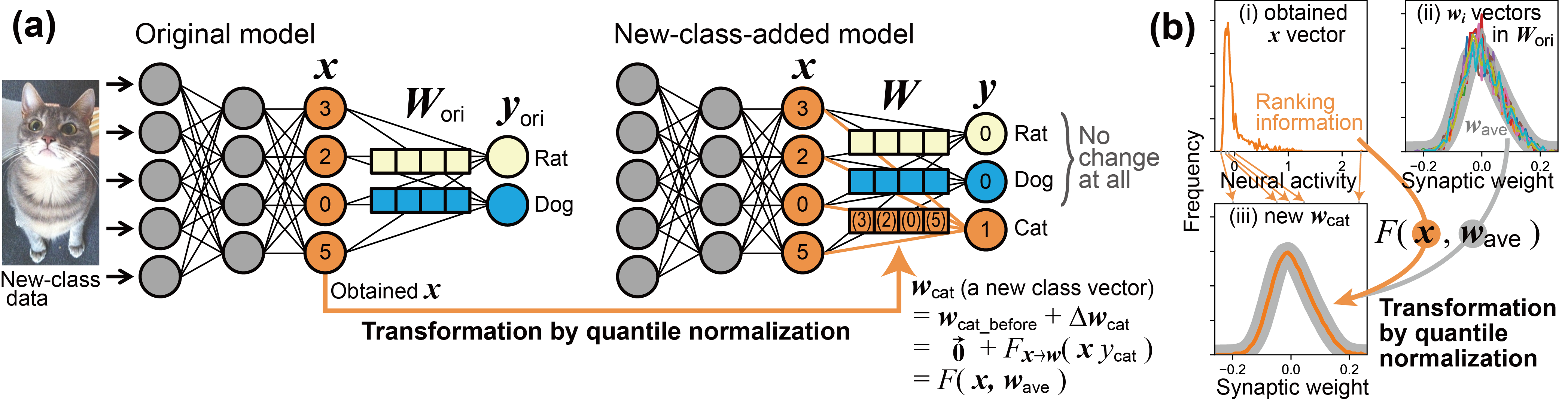}
  \caption{Scheme of DONE. (a) The neural activity input of final dense layer (orange $\bm{x}$ vector in original model) obtained from a new-class data (an image of a cat) is directly used for the transformation to the new-class vector (orange $\bm{w}_{\rm{cat}}$) in the new weight matrix ($\bm{W}$) without any modification to the backbone model. (b) An example case of transformation from $\bm{x}$ to $\bm{w}_{\rm{cat}}$, with actual distribution data when the backbone DNN is EfficientNet-B0. }
  \label{fig:scheme}
\end{figure*}

In this paper, we introduce a very simple method, Direct ONE-shot learning (DONE) with a nonparametric weight imprinting. As shown in Figure~\ref{fig:scheme}(a), DONE directly transform the input of the final dense layer ($\bm{x}$ vector in the figure) obtained by one input image belonging to a new class (e.g., a cat in the figure) into the weight vector for the new additional class ($\bm{w}_{\rm{cat}}$, a row vector of the weight matrix $\bm{W}$). Then, it is done. DONE uses weight imprinting but never modifies backbone DNN including original weight matrix $\bm{W}_{\rm{ori}}$ unlike Qi's method. Qi's method was inspired by the context of metric learning, but DONE was inspired by Hebbian theory \cite{hebb_organization_2002}. This difference in inspiration source makes an important performance difference with small procedure differences.

We here explain the process of our method as a weight imprinting.
In weight imprinting, we can assume that the new weight vector $\bm{w}_{\rm{cat}}$ is born out of nothing and thus is equal to its change, i.e., $\bm{w}_{\rm{cat}} = \vec{0} + \Delta\bm{w}_{\rm{cat}} = \Delta\bm{w}_{\rm{cat}}$. 
Hebbian theory is about this $\Delta\bm{w}_{\rm{cat}}$ and states that a synaptic weight is strengthened when both its presynaptic and postsynaptic neurons are active simultaneously. When a single image of a new class (cat) is presented as visual input, some of the presynaptic neurons  $\bm{x}$ become active. Simultaneously, a postsynaptic neuron corresponding to cat is active (e.g., $y_{\rm{cat}}=1$), while postsynaptic neurons for all the $i$-th original classes are not ($y_{i}=0$), because the training image is known to be a cat. In a general form of Hebbian rule \cite{dayan_theoretical_2001}, the change in the weight vector can be described as $\Delta\bm{w}_{\rm{cat}} \propto \bm{x} \cdot y_{\rm{cat}}$, thus $\bm{w}_{\rm{cat}} = \Delta\bm{w}_{\rm{cat}} \propto \bm{x}$, while $\Delta\bm{w}_{i}=0$ because $y_{i}=0$.

Here, a problem arises with general forms of Hebbian rule alone, because neural activity and synaptic weight are different in scale and those relationships would not be linear, not only in real brains (as physical constraints of neurons) but also in ANNs. For example, Figure~\ref{fig:scheme}(b)-(i) and (b)-(ii) show frequency distributions of neural activity in $\bm{x}$ and weight in $\bm{w}_{i}$, which are different in shape, in an actual DNN. If only the new $\bm{w}_{\rm{cat}}$ had far different statistical properties compared to the other $\bm{w}_{i}$, the comparison between classes would be unequal, and the additional $\bm{w}_{\rm{cat}}$ could inhibit the classification of the original classes (shown later). Therefore, a function for the nonlinear scale transformation is required as an augmentation of Hebbian plasticity, i.e., $\bm{w}_{\rm{cat}} = F_{\bm{x} \rightarrow \bm{w}}( \bm{x} \cdot y_{\rm{cat}} ) $.

DONE takes into account this transformation by quantile normalization \cite{amaratunga_analysis_2001,bolstad_comparison_2003}, so that the frequency distribution of the new $\bm{w}_{\rm{cat}}$ becomes equal to that of $\bm{w}_{\rm{ave}}$ (the average vector of original $\bm{w}_i$ vectors), i.e., $\bm{w}_{\rm{cat}} = F( \bm{x}, \bm{w}_{\rm{ave}} ) $  (Figure~\ref{fig:scheme}(b)-(iii)). 
Quantile normalization is an easy and standard technique in Bioinformatics, and it is suitable for implementing this nonlinear scale transformation. 
The statistical properties of $\bm{w}_{\rm{cat}}$ should be similar to the statistical properties of original synaptic weights.
For example, we could apply linear transformation so that the mean and variance (i.e., 1st and 2nd central moments) of the elements of $\bm{w}_{\rm{cat}}$ are the same as those of $\bm{w}_{\rm{ave}}$. However, it is not clear if such adjustment for only 1st and 2nd central moments is enough in this situation where the 3rd or higher central moments could be different. One of the simplest solutions for every situation is to make all the statistical properties identical, which quantile normalization accomplishes. We call this nonparametric transformation \textit{quantile weight imprinting} (see Methodology section for details).

Our method's basis and procedure are very simple, but it achieves similar accuracy to Qi's method and avoids a severe problem of Qi's method. The severe problem is strong interference with the original classification depending on backbone DNNs (see below, e.g., Figure~\ref{fig3}(e)-(ii)), which would be the reason why weight imprinting has not been widely used yet despite its excellent performance.
DONE achieved over 50\% accuracy (approximately 80\% of original classification) in a one-shot image classification task that adds eight new classes to a model pretrained for the ImageNet 1000 classes (ViT (Vision Transformer) \cite{dosovitskiy_image_2020} or EfficientNet \cite{tan_efficientnet_2019}) as a backbone model (note that the chance level is less than 0.1\%). In a typical five-way one-shot classification task, DONE with ViT achieved over 80\% accuracy.

The advantages of DONE over other weight imprinting methods are (i) simpler basis and procedure with no modification to backbone models, and (ii) nonparametric procedure for little interference with the original classification. The advantages of DONE as a weight imprinting are (iii) no optimization thus little calculation cost and (iv) no parameters or hyperparameters thus reproducible for anyone. 
In addition to proposing the new methods, this paper contains the following useful information: a generic task to add new classes to 1000-class ImageNet models, and the difference in backbone DNNs, specifically, between a Transformer (ViT) and CNNs.

\section{Related work}

\subsection{One-shot and few-shot image classification}

Typical learning approaches for one- or few-shot image classification are metric learning, data augmentation, and meta learning. Weight imprinting has come out from metric learning. Each of these approaches has its own advantages and purposes, and they are not contradictory but can be used in a mixed manner. 

Metric learning uses a classification at a feature space as a metric space \cite{weinberger_distance_2009,snell_prototypical_2017,kaya_deep_2019}. Roughly speaking, metric learning aims to decrease the distances between training data belonging to the same class and increase the distance between the data belonging to different classes. Metric learning such as using Siamese network \cite{koch_siamese_2015} is useful for tasks that require one-shot learning, e.g., face recognition. A Data-augmentation approach generatively increases the number of training inputs \cite{vinyals_matching_2016,schwartz_delta-encoder_2018}. This approach includes various types such as semi-supervised approaches and example generation using Generative adversarial networks \cite{goodfellow_generative_2014}. Meta learning approaches train the abilities of learning systems to learn \cite{andrychowicz_learning_2016,finn_model-agnostic_2017}. The purpose of meta learning is to aim to increase the learning efficiency itself, and this is a powerful approach for learning from a small amount of training data, typically one-shot learning task \cite{huisman_survey_2021}. 

\subsection{Weight imprinting}

Weight imprinting is a learning method that initially arose from an innovative idea ``learning without optimization'' \cite{qi_low-shot_2018}, and DONE is a type of weight imprinting. Weight imprinting does not contain any optimization algorithm and is basically inferior to other optimization methods by themselves in accuracy. However, comparisons of weight imprinting methods with other optimization methods are useful in evaluating those optimization methods, because the performance of weight imprinting methods is uniquely determined by the backbone DNN without any randomness. Thus, weight imprinting does not aim for the highest accuracy in transfer learning or learning from scratch, but for practical convenience in class addition tasks and for reference role as a baseline method.

We here explain the basis of weight imprinting and then specific procedure of Qi's method. Let us consider the classification at the final dense layer of DNN models in general. In most cases, the output vector $\bm{y}=(y_1,\cdots,y_N)$ of the final dense layer denotes the degree to which an image belongs to each class and is calculated from the input vector of the final dense layer $\bm{x}=(x_1,\cdots,x_M)$, weight matrix $\bm{W}$ ($N \times M$), and bias vector $\bm{b}=(b_1,\cdots,b_N)$. Here, for $i$-th class in $N$ classes ($i=1,2,\cdots,N$), a scalar $y_i$ is calculated from the corresponding weight vector $\bm{w}_i=(w_{i 1},\cdots,w_{i M})$ ($i$-th row vector of $\bm{W}$ matrix) and bias scalar $b_i$ as the following equation: 
\begin{equation}
y_i = \bm{x} \cdot \bm{w}_i + b_i = ||\bm{x}||_2 \, ||\bm{w}_i||_2 \cos\theta +b_i,
\end{equation}
where the cosine similarity is a metric that represents how similar the two vectors $\bm{x}$ and $\bm{w}_i$ are irrespective of their size. Thus, this type of model contains cosine similarity as a part of its objective function. 

Weight imprinting uses this basis of the cosine similarity. The cosine similarity will have the maximum value $1$ if $\bm{x}$ and $\bm{w}_i$ are directly proportional. Thus, if a certain $\bm{x}$ is directly used for the weight of a new $j$-th class $\bm{w}_j$ ($j=N+1,\cdots$), the cosine similarity for $j$-th class becomes large when another $\bm{x}$ with a similar value comes.

In Qi's method, to focus only on the cosine similarity as a metric for the objective function, the backbone DNN models are modified in the following three parts:
\begin{itemize}
  \item $\mathbf{Modification\,1:}$ Adding $L_2$ normalization layer before the final dense layer so that $\bm{x}$ becomes unit vector, i.e., $||\bm{x}||_2=1$
  \item $\mathbf{Modification\,2:}$ Normalizing all $\bm{w}_i$ to become unit vectors, i.e., $||\bm{w}_i||_2=1$ for all $i$.
  \item $\mathbf{Modification\,3:}$ Ignoring all bias values $b_i$, i.e., $\bm{b}$ vector.
\end{itemize}
Then, the final-dense-layer input obtained from a new-class image $\bm{x}_{\rm{new}}$ ($L_2$-normalized, in Qi's method) is used as the weight vector for the new class $\bm{w}_j$, i.e.,
\begin{equation}
    \bm{w}_j = \bm{x}_{\rm{new}}.
\end{equation}

Qi's method is already simple and elegant, but it still requires some modifications to the backbone DNN, which involves changes in the objective function. Whether a modification is good or bad depends on the situation, but if not necessary, it would be better without modification in order to avoid unnecessary complications and unexpected interference with the original classification because the backbone DNN would be already well optimized for a certain function. Also, Qi's method uses linear transformation for conversion of $\bm{x}$ into $\bm{w}_j$ as a result of focusing on the cosine similarity, without considering the difference in statistical properties between $\bm{x}$ and $\bm{w}_j$, which limits the range of backbone DNNs used. There have been various researches that make Qi's method more complex and applicable \cite{dhillon_baseline_2019,passalis_hypersphere-based_2021,li_classification_2021,cristovao_few_2022,zhu_weight_2022}, but to the best of our knowledge, none that make it simpler or solve the transformation problem.

\section{Methodology}

\subsection{Procedure and basis of DONE}

DONE does not modify backbone DNN and just directly applies $\bm{x}_{\rm{new}}$ to $\bm{w}_j$ ($j=N+1,\cdots$), as shown in Figure~\ref{fig:scheme}, as
\begin{eqnarray}
    & \bm{w}_j = F(\bm{x}_{\rm{new}},\bm{w}_{\rm{ave}}),&\\ 
    & b_j = \bm{\tilde{b}}_{\rm{ori}}, & 
\end{eqnarray}
where $F(\bm{x}_{\rm{new}},\bm{w}_{\rm{ave}})$ is a quantile normalization of $\bm{x}_{\rm{new}}$, using the information of the average weight vector for original classes ($\bm{w}_{\rm{ave}}$) as the reference distribution, and $\bm{\tilde{b}}_{\rm{ori}}$ is the median of the original bias vector $\bm{b}_{\rm{ori}}$. Then, it is done. 

In the quantile normalization, the elements value of resultant $\bm{w}_j$ become the same as that of the reference $\bm{w}_{\rm{ave}}$. Specifically, for example, first we change the value of the most (1st) active neuron in $\bm{x}_{\rm{new}}$ into the highest (1st) weight value in $\bm{w}_{\rm{ave}}$. We then apply the same procedure for the 2nd, 3rd, $\cdots,M$-th highest neurons. Then, the ranking of each neuron in $\bm{x}_{\rm{new}}$ remains the same, and the value of each ranking is all identical to $\bm{w}_{\rm{ave}}$. This resultant vector is $\bm{w}_j$. Namely, all statistical properties of the elements of $\bm{w}_j$ and $\bm{w}_{\rm{ave}}$ are identical (frequency distributions are the same). 

For $\bm{w}_{\rm{ave}}$, to implement the concept of physical constraints of neurons, we used all the $N \times M$ elements of flattened $\bm{W}_{\rm{ori}}$, divided these elements into $M$ parts in ranking order, and obtained the $M$ median values in each $N$ element as $M$-dimensional $\bm{w}_{\rm{ave}}$. For example, in the case of ViT-B/32 ($N=1000$, $M=768$), the highest value of $\bm{w}_{\rm{ave}}$ is the median of 1st to 1000th highest elements in 768,000 elements of $\bm{W}_{\rm{ori}}$, and the lowest value is the median of 767,001th to 768,000th elements. (See Figure S1 for comparison with another averaging method.)

\subsection{Limitations, applications, and negative impacts} 

As limitations, DONE requires a neural network model that has a dense layer for classification as above. DONE can be used for a wide range of applications with DNN classifiers, including OOD applications\cite{yang_generalized_2021}. There can be various potential negative societal impacts associated with these broad applications. One example is immoral classification or discrimination when classifying human-related data, such as facial images, voices, and personal feature data.

\subsection{Implementation and dataset}

As backbone models, we employed 
ViT-B/32 \cite{dosovitskiy_image_2020}
with ``vit-keras'' \cite{morales_vit-keras_nodate},
EfficientNet-B0 \cite{tan_efficientnet_2019}
with ``EfficientNet Keras (and TensorFlow Keras)'' \cite{iakubovskii_efficientnet_nodate}, 
InceptionV1 \cite{szegedy_going_2015} (employed in \citet{qi_low-shot_2018})
with ``Trained image classification models for Keras'' \cite{andrews_trained_nodate},
ResNet-12 \cite{he_deep_2016}
with ``tf2cv'' \cite{semery_tf2cv_nodate}, and 
ResNet-50 \cite{he_deep_2016},
MobileNetV2 \cite{sandler_mobilenetv2_2018},
VGG-16 \cite{simonyan_very_2015}
with Tensorflow \cite{abadi_tensorflow_2015}.
All models used in this study were pretrained with ImageNet. 

We used CIFAR-100 and CIFAR-10 \cite{krizhevsky_learning_2009} for additional classes with Tensorflow \cite{abadi_tensorflow_2015}. Also, for transfer learning, we used CIFAR-FS \cite{bertinetto_meta-learning_2018} with Torchmeta \cite{deleu_torchmeta_2019}. We used ImageNet (ILSVRC2012) images \cite{deng_imagenet_2009,russakovsky_imagenet_2015} for testing the performance of the models. We used information of 67 categorization \cite{eshed_novelty_2020} of ImageNet 1000 classes, for a coarse 10 categorization in Figure~\ref{fig4}(a).

\begin{figure*}[t]
  \centering
  \includegraphics[width=0.95\textwidth]{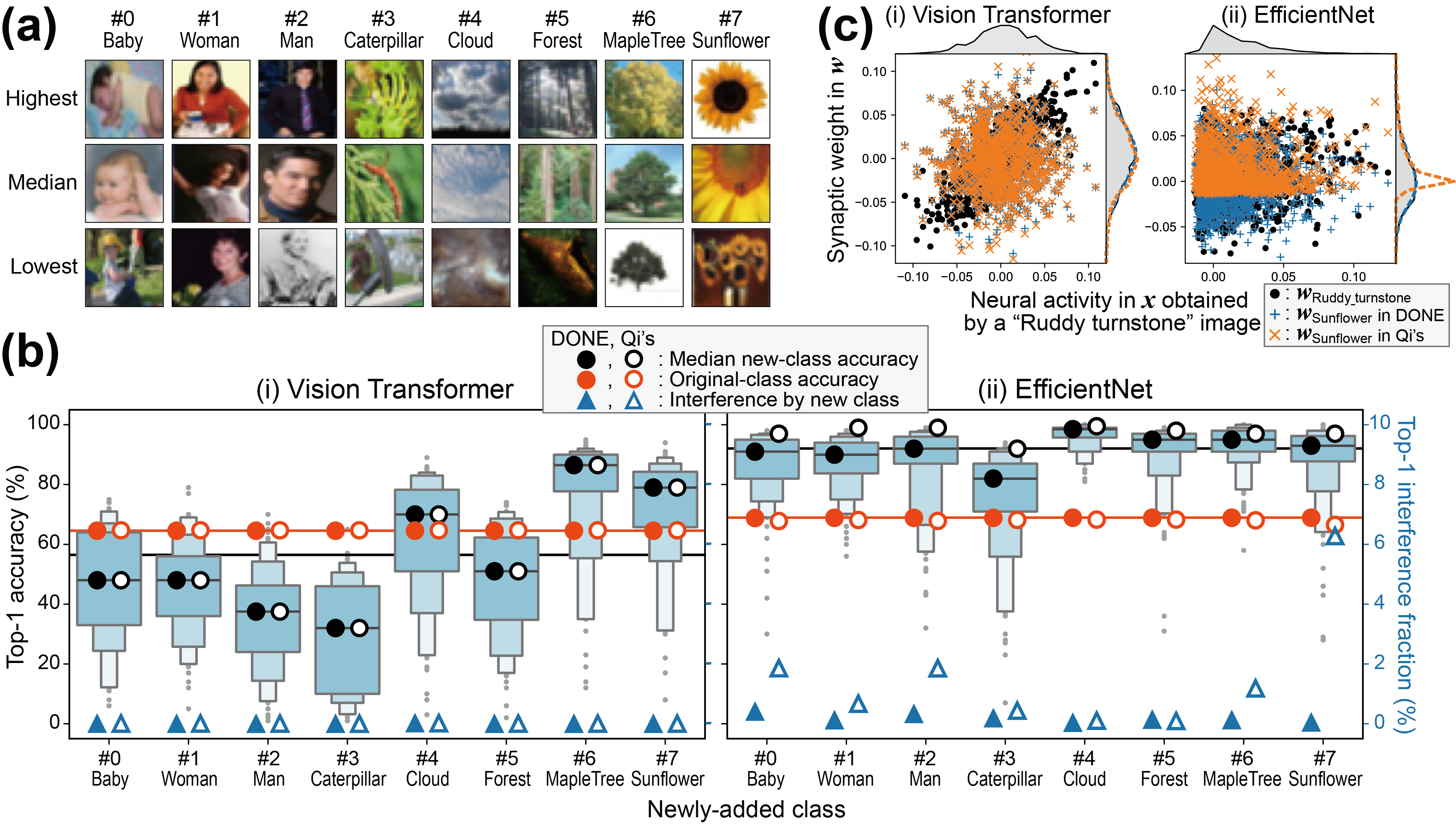}
  \caption{One-class addition by one-shot learning.
  (a) Representative images of the newly-added CIFAR-100 classes. Each image was chosen as a representative because the model that learned the image showed the highest, median, and lowest accuracy in each class in (b)-(i). (b) Letter-value plots of top-1 accuracy of the one-class-added models obtained by one-shot learning with DONE in classification of new-class images. The median top-1 accuracy of the new-class classification (black circles), top-1 accuracy in original-class classification (orange circles), and the fraction of the interference with the original-class classification by the newly-added class (blue) are also plotted for DONE (closed) and Qi's method (open). Black and orange lines show the mean of the 8  closed circles. (c) The relationship between $\bm{x}$ and $\bm{w}$ vectors when an image of ``Ruddy turnstone'' is input and it is miss-classified as ``Sunflower'' only in the case of Qi's method with EfficientNet. The frequency distributions of elements of each vector are also shown outside the plot frames. }
  \label{fig2}
\end{figure*}

\section{Results and Discussion}

\subsection{One-class addition by one-shot learning}

First, according to our motivation, we investigated the performance of DONE when a new class from one image was added to a DNN model pretrained with ImageNet (1000 classes). As new additional classes, we chose eight classes, ``baby'', ``woman'', ``man'', ``caterpillar'', ``cloud'', ``forest'', ``maple\_tree'', and ``sunflower'' from CIFAR-100, which were not in ImageNet (shown in Figure~\ref{fig2}(a)). The weight parameters for the additional one class $\bm{w}_j$ is generated from one image, thus the model had 1001 classes. To conduct stochastic evaluations, 100 different models were built by using 100 different training images for each additional class.

Figure~\ref{fig2}(b) shows letter-value plots of the accuracy for each additional class (chance level $1/1001$). The mean of the median top-1 accuracy of 8 classes by DONE were 56.5\% and 92.1\% for ViT and EfficientNet, respectively (black line). When the mean accuracy was compared with the accuracy of ImageNet validation test by the original 1000-class model (orange line; 65\% and 69\%), the mean with ViT had no significant difference and the mean of EfficientNet was significantly greater (one sample $t$-test; two-sided with $\alpha$=0.05, in all statistical tests in this study). The higher accuracy than the original classes in EfficientNet is strange, and it is considered that EfficientNet tends to recognize the new-class images as just OOD (see later, Figure~\ref{fig4}).

An obvious fact in one-shot learning is that a bad training image produces a bad performance, e.g., the accuracy was 6\% in ViT when the training image was a baby image shown at the bottom left in Figure~\ref{fig2}(a). But in practical usage, a user is supposed to use a representative image for the training. We therefore think that the low performance due to a bad training image is not a significant issue. 

We investigated the interference of the class addition with the classification performance of the original 1000 classes. 
We evaluated the original 1000-class model and eight 1001-class models that showed the median accuracy, by using all 50,000 ImageNet validation images (Figure~\ref{fig2}(b)). 
The difference between the accuracy of the original 1000-class model (orange line) and the mean accuracy of the eight 1001-class models (orange closed circles) was less than 1\% (0.004\% and 0.664\% for ViT and EfficientNet, respectively). 

Figure~\ref{fig2}(b) also shows the fraction of ImageNet validation images in which the output top-1 answer of the added model was the new class (thus incorrect) in the 50,000 images (blue closed triangles; right axis). This interference fraction was low in ViT, and for example, only 2 images out of 50,000 were classified as ``baby''. When we checked the two images, both images indeed contained a baby though its class in ImageNet was ``Bathtub''. Therefore, observed interference in ViT was not a mistake but just the result of another classification. 
EfficientNet shows a significantly greater fraction of interference than ViT (Wilcoxon signed-rank test), but we also confirmed that a similar thing happened, e.g., 198 of the 204 ImageNet-validation images classified as ``baby'' in EfficientNet contained human or doll.

We also compared DONE with Qi's method. Open circles and triangles show the results using Qi's method instead of DONE in the same tests described above. When the backbone model was EfficientNet, the strangely-high accuracy (paired sample $t$-test) and the interference fraction (Wilcoxon signed-rank test) were significantly greater by Qi's method than by DONE. Also, a significant outlier of decreased accuracy in the ImageNet validation test was observed (orange open circle for ``Sunflower''; Smirnov-Grubbs test). On the other hand, those differences were not significant in the case of ViT.

To investigate the cause of the difference between DONE and Qi's method, especially about the greater interference by Qi's method in EfficientNet, we plotted $\bm{w}_{\rm{Sunflower}}$ and $\bm{w}_{\rm{Ruddy\_turnstone}}$ against $\bm{x}$ obtained from an image of ``Ruddy turnstone'' (Figure~\ref{fig2}(c)). Note that all the vectors here are $L_2$-normalized, and thus DONE and Qi's method have common $\bm{w}_{\rm{Ruddy\_turnstone}}$ and $\bm{x}$. In the case of ViT, the shape of the frequency distributions of all these vectors are similar, and $\bm{w}_{\rm{Sunflower}}$ of DONE and Qi's method are similar.

On the other hand, in EfficientNet, the shape of frequency distributions are more different between $\bm{w}_{\rm{Ruddy\_turnstone}}$ and $\bm{x}$ than ViT, and thus the shape of frequency distributions are more different between $\bm{w}_{\rm{Ruddy\_turnstone}}$ and $\bm{w}_{\rm{Sunflower}}$ by Qi's method than by DONE. Then, by Qi's method, $\bm{x}$ is more similar to $\bm{w}_{\rm{Sunflower}}$ than $\bm{w}_{\rm{Ruddy\_turnstone}}$ because not neural match but statistical properties are similar, i.e., Qi's method with EfficientNet tends to classify every image into the new class. This is the basis of the problem by the linear transformation of neural activity to synaptic weight. Therefore, the difference between DONE and Qi's method appears in the interference when the statistical properties of $\bm{x}$ and $\bm{w}_i$ vectors in the backbone DNN are different (thus the results in ViT are similar between DONE and Qi's method).

\begin{figure*}[t]
  \centering
  \includegraphics[width=0.95\textwidth]{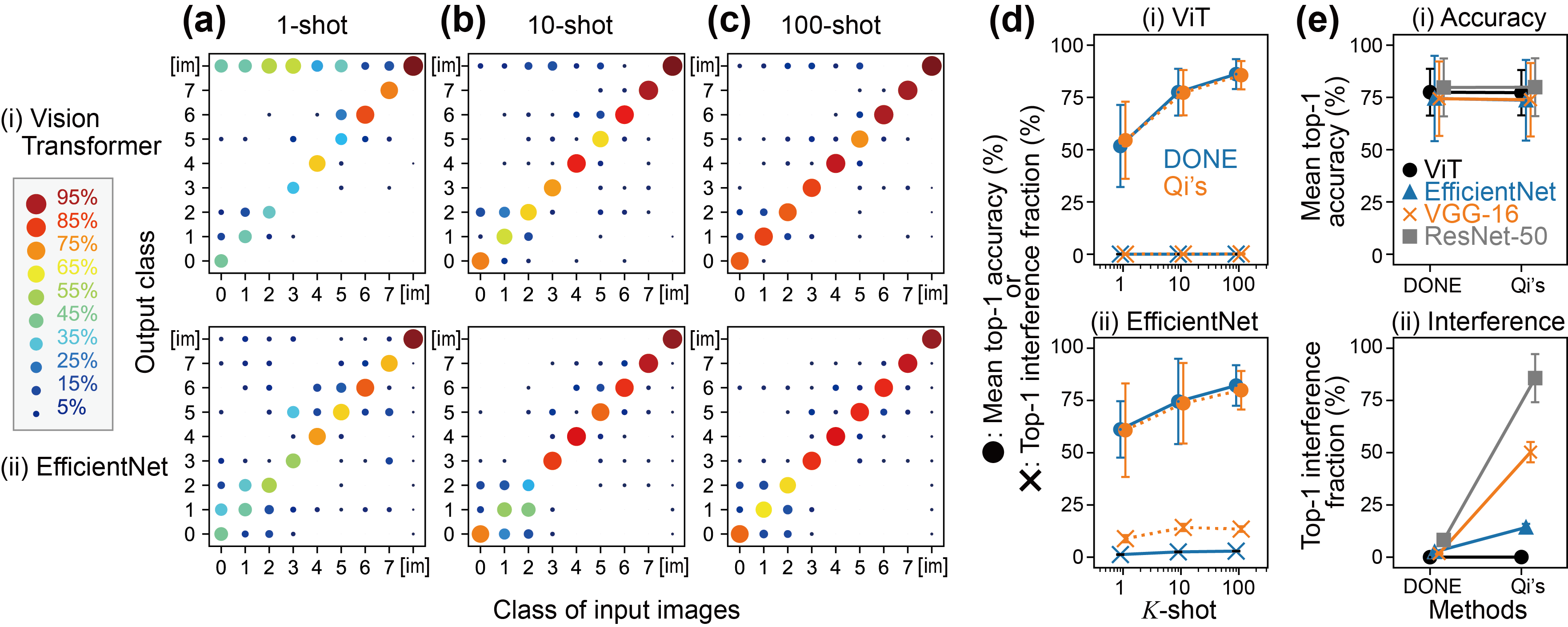}
  \caption{Multi-class addition and $K$-shot learning. (a), (b), and (c) show the results of the 1008-class model constructed by 1, 10, and 100-shot learning, respectively. The horizontal and vertical axes show the class of the input images, and the output class, respectively. The class numbers are those shown in Figure~\ref{fig2}(a). The class [im] contains 1000 classes of ImageNet. (d) Summary of the mean accuracy and the interference with original-class classification by DONE and Qi's method. (e) Results of the 1008-class model constructed by 10-shot learning with 4 different backbone models. Error bars show SD of the 8 classes.  }
  \label{fig3}
\end{figure*}

\subsection{Multi-class addition and K-shot learning}

DONE was able to add a new class as above, but it might just be because the models recognized the new-class images as OOD, i.e., something else. Therefore, it is necessary to add multiple new similar classes and check the classification among them. In addition, it is necessary to confirm whether the accuracy increases by increasing the number of training images, because in practical uses, users will prepare not just one training data but multiple training data for each class. 

Specifically, we used one image from each of the eight classes and added new eight classes to the original 1000 classes, using DONE as one-shot learning. We evaluated this 1008-class model by 100 CIFAR test images for each of 8 classes and 10,000 ImageNet validation images. Figure~\ref{fig3}(a) shows the results of the output of the representative model constructed by one-shot learning in which one image that showed median accuracy in Figure~\ref{fig2}(b) was used as a standard training image of each class. In both backbone DNNs, the fraction of output of the correct class was the highest among the 1008 classes, and mean top-1 accuracy of the 8 classes was 51.8\% and 61.1\% in ViT and EfficientNet, respectively. That is, DONE was also able to classify newly added similar classes together with the original classes, in both DNNs.

Next, we increased the number of training images as $K$-shot learning. In the case of 10-shot learning (Figure~\ref{fig3}(b)), each of the ten images was input to obtain each $\bm{x}$, and the mean vector of the ten $\bm{x}$ vectors was converted into $\bm{w}_j$, according to the Qi's method. For this representative 10-shot model, we used 10 images whose index in CIFAR-100 was from the front to the 10th in each class. We also tested 100-shot learning in the same way (Figure~\ref{fig3}(c)). As a result, we found that such a simple averaging operation steadily improved the accuracy (Figure~\ref{fig3}(d) summaries the mean accuracy).

When we used Qi's method, compared to the case of DONE, ImageNet images were significantly more often categorized to the new classes as interference when the backbone model was EfficientNet (paired sample $t$-test), while there was no significant difference in the mean accuracy between DONE and Qi's method with both backbone DNNs (Figure~\ref{fig3}(d)). 

Figure~\ref{fig3}(e) shows the results of the 10-shot learning using various well-known DNN models. While accuracy was similar in those two methods, Qi's method showed severe interference that more than half of original-class ImageNet images were classified into new classes (50\% and 86\% in VGG-16 and ResNet-50, respectively) despite the interference chance level of 8/1008, and DONE reduced the interference to less than one tenth (1.8\% and 8.0\%, respectively; DONE reduced the interference to one fourth even in ViT). Thus, DONE can avoid the DNN-dependent severe interference in Qi's method. 
The cause of this interference is considered to be the difference in the shape of the frequency distributions of $\bm{x}$ and $\bm{w}_{i}$ (right-tailed and bell-shaped in most CNNs). Since we have observed this difference in distribution in all six CNNs we have examined (Figure S2), it is likely that most CNNs cause this interference, which DONE can avoid.

\begin{figure*}[t]
  \centering
  \includegraphics[width=0.95\textwidth]{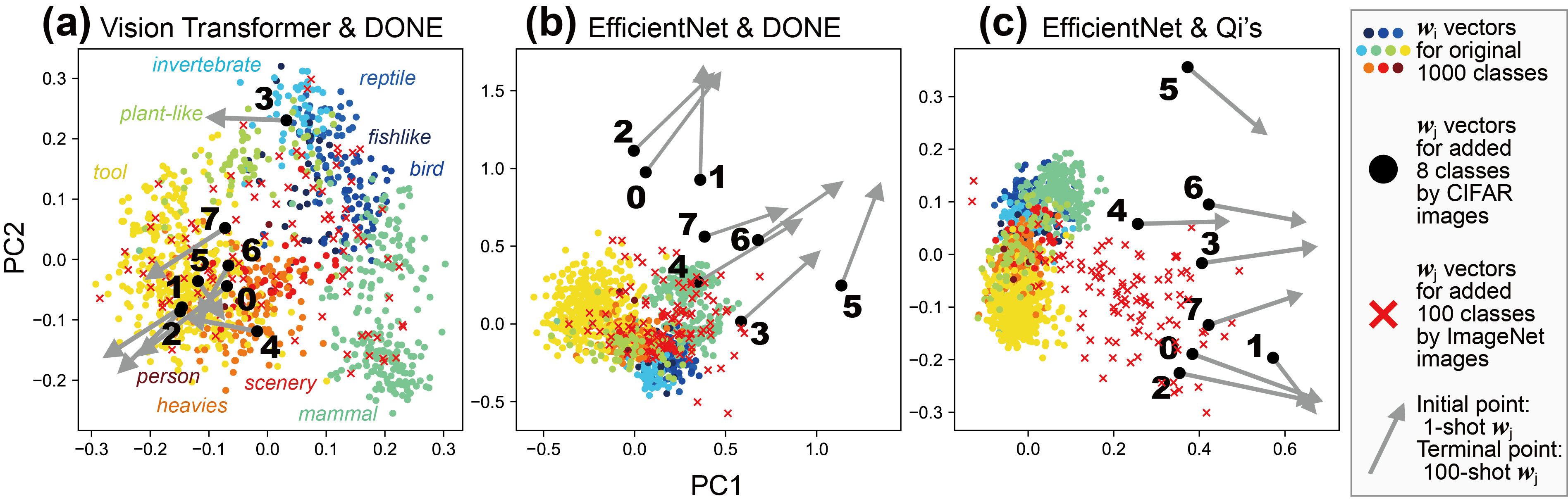}
  \caption{Principal component analysis of weight vectors. PCA of each $\bm{w}_i$ and $\bm{w}_j$ vector in the one-shot 1008-class models shown in Figure~\ref{fig3}(a). Different colors for $\bm{w}_i$ show a coarse 10 categorization of the classes. Also, 100 $\bm{w}_j$ vectors obtained by inputting 100 ImageNet images are shown. }
  \label{fig4}
\end{figure*}

\subsection{Principal component analysis of weight vectors}

Qi's method showed greater interference in classification of the original-class images than DONE when the backbone DNN is EfficientNet. Moreover, even by DONE, EfficientNet showed greater interference than ViT and strangely-high accuracy at 1001-class model, even though DONE did not change the weights for the original classes and transformed the new-class weights so that the statistical properties were the same as those of the original-class weights. Therefore, there should be at least two reasons for these results shown in EfficientNet.
To investigate those reasons, we analyzed $\bm{W}$ matrix ($\bm{w}_i$ and $\bm{w}_j$ vectors) of the one-shot 1008-class models shown in Figure~\ref{fig3}(a) (and corresponding models by Qi's method) by Principal component analysis (PCA; Figure~\ref{fig4}). 

In ViT by DONE (Figure~\ref{fig4}(a); Qi's methods showed similar results, see Figure S3), newly added $\bm{w}_j$ vectors (black circles, with the ID number of newly-added 8 classes) were comparable to those of the original classes $\bm{w}_i$ (colored circles), e.g., $\bm{w}_j$ vector of a new class ``caterpillar (\textsf{3} in Figure~\ref{fig4}(a))'' was near $\bm{w}_i$ of original ``invertebrate'' classes. Also, even when we got $\bm{w}_j$ by inputting ImageNet images (red crosses; validation ID from the front to the 100th), those ImageNet $\bm{w}_j$ vectors distributed in similar range.

On the other hand, in EfficientNet by DONE (Figure~\ref{fig4}(b)), most of newly-added 8-class $\bm{w}_j$ were out of the distribution (meaning out of minimal bounding ellipsoid) of $\bm{w}_i$ of original 1000 classes, while most of the ImageNet $\bm{w}_j$ (red crosses) were inside the distribution of $\bm{w}_i$. 
Therefore, in the case of DONE, the main reason for the observed greater interference and strangely-high accuracy in EfficientNet than ViT would be the difference between ImageNet and CIFAR. 
These results are consistent with known facts that ViT is considered to be better at predictive uncertainty estimation \cite{guo_calibration_2017,minderer_revisiting_2021}, more robust to input perturbations \cite{bhojanapalli_understanding_2021}, and more suitable at classifying OODs \cite{fort_exploring_2021} than CNNs like EfficientNet. 

In EfficientNet by Qi's method (Figure~\ref{fig4}(c)), most of not only 8-class $\bm{w}_j$ but also the ImageNet $\bm{w}_j$ were out of the distribution of $\bm{w}_i$ of original 1000 classes. The difference in the distributions between the original $\bm{w}_i$ and the ImageNet $\bm{w}_j$ is considered to indicate the difference in the statistical properties of $\bm{x}$ and $\bm{w}_i$ vectors in EfficientNet.

In the case of 100-shot learning (the terminal points of the gray arrows in Figure~\ref{fig4}), $\bm{w}_j$ went away from the cluster of original $\bm{w}_i$ in all three cases, although their performance was better than one-shot learning. Therefore, 100-shot $\bm{w}_j$ were considered to work somehow in a different way from the original $\bm{w}_i$.

\begin{figure*}[t]
  \centering
  \includegraphics[width=0.95\textwidth]{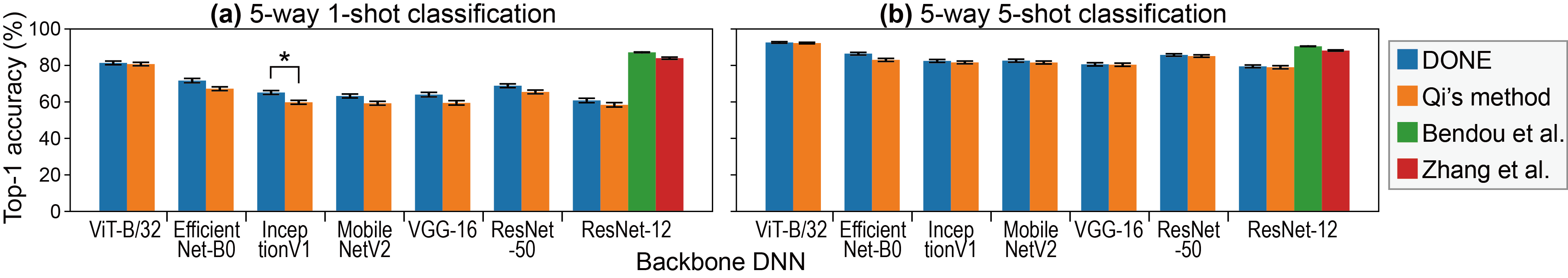}
  \caption{5-way 1-shot (a) and 5-shot (b) classification accuracy on CIFAR-FS with various backbone DNNs. Error bars show standard errors. Asterisks mean significant differences between DONE and Qi's method (Dwass-Steel-Critchlow-Fligner test). }
  \label{fig5}
\end{figure*}

\subsection{Transfer few-shot learning}

DONE is recommended for the easy addition of new classes, not for transfer learning. However, DONE can work for transfer learning (Figure S4) and is convenient for the evaluation of DNNs and other few-shot learning methods. We examined the 5-way (5 classes) 1-shot task of CIFAR-FS, which is a kind of standard task in one-shot classification. Specifically, we used each single image in 5 classes out of 100 classes of CIFAR-100 for constructing a model, and evaluated the model by 15 images in each class. The combination of the 5 classes (and corresponding training images) was randomly changed 100 times (Figure~\ref{fig5}(a)). Also 5-way 5-shot tasks were tested in a similar way (Figure~\ref{fig5}(b)).

We found ViT significantly outperformed the other DNNs in all conditions (Dwass-Steel-Critchlow-Fligner test). 
DONE was significantly better than Qi's method with a CNN model and never significantly worse.

Figure~\ref{fig5} also clearly shows that how much other state-of-the-art one-shot learning methods with optimization (methods in \cite{zhang_sample-centric_2022} and \cite{bendou_easy_2022}) outperform DONE as the baseline without optimization, at the same test with a common backbone DNN (ResNet-12).
Thus, using a better backbone DNN in the immediate use of DONE without optimization has a similar effect to running optimization.

\section{Conclusion and Future work}

This paper has proposed DONE, one of the simplest one-shot learning methods, which avoids the severe problem of the previous weight imprinting method and can add new classes to a pretrained DNN at a decent performance without optimization or modification of the DNN.
DONE applies quantile weight imprinting, which is an implementation of the concept of physical constraints of neurons, to the final dense layer of a DNN model. Given the simplicity and wide applicability, not only DONE but also quantile weight imprinting alone are expected to be applied in a wide range of the field of ANN. 
This study has just proposed the method, and its scalability (Figure S5) and expected applications are yet to be elucidated. 
Since the performance of DONE is completely dependent on backbone DNNs and further development of DNN is certain, there will be more and more tasks that can be done by DONE.
Moreover, DONE may provide useful insights into the brain's learning principles because of the simple and Hebbian-like basis.

\section{Acknowledgments}
We thank Dr. Toshio Yanagida, Dr. Toshiyuki Kanoh, Dr. Kunihiko Kaneko, Dr. Tsutomu Murata, Dr. Tetsuya Shimokawa and Mr. Yosuke Shinya for helpful advice and discussions. This work was supported in part by MIC under a grant entitled ``R\&D of ICT Priority Technology (JPMI00316)'' and JSPS KAKENHI Grant Number JP20H05533.


\bibliography{aaai23.bib}

\end{document}